\documentclass[runningheads]{llncs}
\usepackage[T1]{fontenc}
\usepackage{graphicx}
\usepackage{graphicx}
\usepackage{amssymb}
\usepackage{amsmath}
\usepackage{multirow}
\usepackage{xcolor}

\usepackage[colorlinks=true, allcolors=blue]{hyperref}

\begin{document}
\title{Kinetic Mining in Context: Few-Shot Action Synthesis via Text-to-Motion Distillation}
\titlerunning{Kinetic Mining in Context}
%
\author{
Luca Cazzola\inst{1}\orcidID{0009-0000-6285-8342} \and
Ahed Alboody\inst{2}\orcidID{0000-0002-9555-7471}
}
\authorrunning{L. Cazzola et al.}
%
\institute{
University of Trento, Via Sommarive 5, 38123 Trento, Italy \\ \email{luca.cazzola-1@studenti.unitn.it} \and
CESI LINEACT, 13 Avenue Simone Veil, 06200 Nice, France \\ \email{aalboody@cesi.fr}}
\maketitle              

\begin{abstract}
The acquisition cost for large, annotated motion datasets is a bottleneck for skeletal-based Human Activity Recognition (HAR). While Text-to-Motion (T2M) models offer a compelling, scalable source of synthetic data, their training objectives, which emphasize general artistic motion, and dataset structures fundamentally differ from HAR’s requirements for kinematically precise, class discriminative actions. We propose KineMIC (Kinetic Mining In Context), a transfer learning framework for few-shot action synthesis. KineMIC adapts a T2M diffusion model to an HAR domain by hypothesizing that semantic correspondences in the text encoding space can provide soft supervision for kinematic distillation. Our solution leverages CLIP text embeddings to mine relevant motion from the source data, guiding the fine-tuning of a generalist backbone into a specialized Action-to-Motion generator. Validated on a subset of NTU RGB+D 120 using only 10 samples per class and considering HumanML3D as source T2M dataset, KineMIC generates coherent motions that provide robust synthetic data, delivering a +23.1\% improvement in action recognition accuracy. Animated illustrations and supplementary materials are available at \url{https://lucazzola.github.io/kinemic-page}.
\end{abstract}

%
%

\section{Introduction}

\begin{figure}[!t]
    \centering
    \resizebox{1.0\textwidth}{!}{\includegraphics{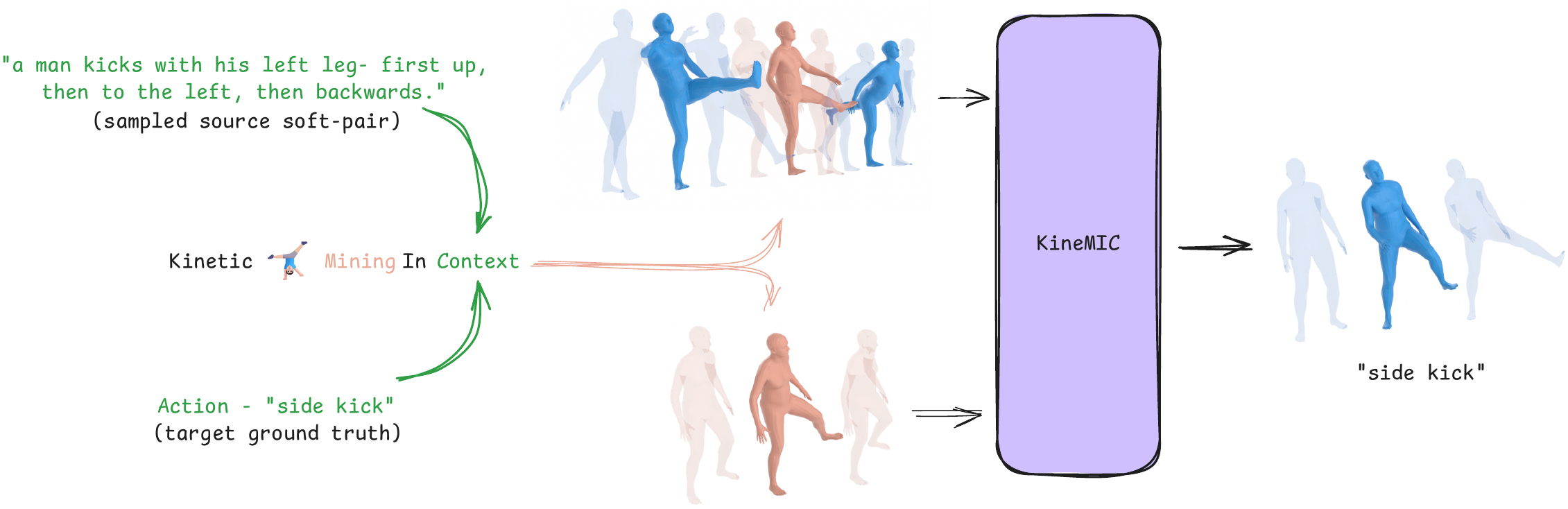}}
    \caption{\textbf{Kinetic Mining in Context}. The target action sample (bottom) is used to contextualize the search within the large source data sample (top), establishing soft pairs. The mining operation identifies a kinematically relevant segment (in orange) from the source data.}
    \label{fig:kinetic_mining_in_context}
\end{figure}

Human Activity Recognition (HAR) has become a cornerstone in a multitude of fields, including sports performance analysis, human-robot collaboration, and intelligent surveillance~\cite{Dentamaro2024HumanAR}. Skeletal-based HAR remains a fundamental modality, frequently employed due to its lightweight representation, robustness to environmental variations, and inherently privacy-preserving nature~\cite{yan2018spatialtemporalgraphconvolutional}. However, the performance of deep learning models for this task is fundamentally limited by the availability of large-scale, accurately annotated datasets. The acquisition and precise labeling of such high-quality, task-specific data is notoriously expensive and labor-intensive, creating a significant bottleneck that hampers progress, particularly in few-shot settings~\cite{Dentamaro2024HumanAR,HungCuong2023DeepLF}. This core challenge of data scarcity in HAR is what we aim to address.

To mitigate this fundamental data bottleneck, much of modern few-shot recognition research focuses on applying strategies such as meta-learning and metric-learning frameworks directly to the action recognition model~\cite{wanyan2025comprehensivereviewfewshotaction}. An alternative approach leverages generative models to create novel synthetic samples, thereby augmenting the small training set~\cite{fukushi2024fewshot}. When it comes to motion synthesis, current research is dominated by Text-to-Motion (T2M) synthesis models, due to the growing interest in text as a conditioning modality. This interest has driven the creation and collection of large-scale T2M datasets~\cite{t2m_humanml,mahmood2019amassarchivemotioncapture}, which has, in turn, led to the development of powerful deep models~\cite{guo2023momaskgenerativemaskedmodeling,t2m_humanml,tevet2022humanmotiondiffusionmodel} that can be employed as general motion priors for other generative tasks~\cite{sawdayee2025dancelikechickenlowrank,shafir2023humanmotiondiffusiongenerative}. T2M priors are strategically appealing as a generative solution because the scalability of their annotation, using free-form, descriptive text, is significantly easier to achieve for general diversity than collecting high-volume, kinematically specific data for HAR. Yet, the challenge of effectively exploiting such general T2M priors for specialized, downstream tasks like HAR remains underexplored.

Our work concentrates on this challenge: adapting a general T2M prior to function as an Action-to-Motion (A2M) synthetic data generator for a specific target HAR domain. This transformation is non-trivial due to a significant domain gap characterized by two key factors. First, a semantic discrepancy exists where source T2M data uses descriptive text, while target HAR requires generation based on discrete action labels. Second, a kinematic gap exists between the broad, fluid motions of the source domain and the short, atomic motions required for HAR. The pre-trained T2M model, being a generalist, is therefore ill-equipped to meet requirements for reliable HAR classification.

To address these challenges and bridge this domain gap for few-shot action synthesis, we propose KineMIC (Kinetic Mining In Context, Fig.~\ref{fig:kinetic_mining_in_context}). Our method employs teacher-student architecture, wherein a frozen teacher, pre-trained on a source T2M dataset, guides the fine-tuning of a student model on the limited target HAR domain. Our method starts by establishing a semantic correspondence between the sparse target action labels and the rich source textual descriptions through CLIP~\cite{radford2021learningtransferablevisualmodels} text encoder. Secondly,  relevant motion sub-sequences, extracted from the vast source dataset, are used to turn the general-purpose student into a specialized generator. The main contributions of this work are as follows:
\begin{itemize}
    \item To the best of the authors' knowledge, we are the first to tackle the challenge of adapting a T2M model into an A2M generator for HAR applications, moreover, addressing this within a few-shot setting.
    \item We introduce KineMIC, a teacher-student framework that specializes a general T2M diffusion prior~\cite{tevet2022humanmotiondiffusionmodel} for domain-specific HAR, yielding significant improvements in recognition accuracy with minimal data.
\end{itemize}

\section{Related Works}

\subsection{Skeletal-based HAR}
Recognizing human activity from skeletal data is a pivotal research area in computer vision. Early approaches modeled motion as time series using RNNs and LSTMs~\cite{du2015hierarchical,ntu60}. A paradigm shift occurred with the introduction of Graph Convolutional Networks (GCNs)~\cite{yan2018spatialtemporalgraphconvolutional}, which re-conceptualized the skeleton as a graph, modelling both spatial and temporal relations. This led to rapid advancements, such as the use of adaptive graph structures~\cite{shi2019twostreamadaptivegraphconvolutional}, refined GCNs~\cite{chi2022infogcn,duan2022pysklgoodpracticesskeleton}, and more recently, specialized architectures like 3D convolutional networks~\cite{duan2022revisitingskeletonbasedactionrecognition} and Transformers~\cite{do2024skateformerskeletaltemporaltransformerhuman,plizzari2021spatio}. Despite these advancements, the field still struggles with data scarcity, as the high capacity of modern models leads to severe overfitting in data-limited regimes. Our work addresses this bottleneck by proposing a novel synthesis framework leveraging deep T2M models, supporting HAR applications in few-shot settings through synthetic data.

\subsection{Generative Models for 3D Skeleton-based Motion}
Realistic human motion synthesis evolved from early VAEs and GANs conditioned on discrete actions~\cite{Guo_2020,petrovich2021actionconditioned3dhumanmotion} to highly expressive models. This progress was fueled by large motion capture datasets~\cite{t2m_humanml,mahmood2019amassarchivemotioncapture} and richer conditioning signals, advancing from text~\cite{petrovich2022temosgeneratingdiversehuman} to even music~\cite{zhang2025motionanythingmotiongeneration} as a modality. The current state-of-the-art is dominated by Denoising Diffusion Probabilistic Models (DDPMs)~\cite{tevet2022humanmotiondiffusionmodel} and masked generative models~\cite{guo2023momaskgenerativemaskedmodeling}, which achieve impressive generative fidelity. However, these T2M models, trained for general character animation, inherently lack the kinematic specificity required for downstream HAR tasks. Our proposed KineMIC framework addresses this domain gap by using a kinetic mining strategy to distill HAR-relevant knowledge from the T2M backbone, enabling its application as a specialized data generator.

\subsection{Few-Shot HAR with Generative Models}
The high cost of annotated motion data bottlenecks Few-Shot Human Activity Recognition (FSHAR). While classifier-side methods (e.g., metric/meta-learning) \cite{wanyan2025comprehensivereviewfewshotaction} are the most popular approach, they are inherently constrained by the limited kinematic diversity of the few-shot support set. A less-explored alternative is deep generative data augmentation \cite{leng2025scalinghumanactivityrecognition,lupion2024dataaugmentation}. Most notably, recently Fukushi et al. \cite{fukushi2024fewshot} demonstrated this using a GAN with cross-domain regularization. To bypass GAN notorious training instability and explore a new avenue, we propose leveraging modern T2M diffusion models for few-shot motion synthesis. In direct contrast to \cite{fukushi2024fewshot}, our approach employs a "semantics-first" matching strategy. By utilizing CLIP \cite{radford2021learningtransferablevisualmodels} text encoding space, we disentangle the synthesis process from the kinematic limitations of the few-shot set.
\section{Problem Formulation}
The core challenge we address is the adaptation of a pre-trained T2M generative model for data augmentation within the context of FSHAR. Our goal is to leverage the extensive kinematic knowledge contained in a large source (i.e. prior) domain to synthesize a high volume of diverse, class-specific motion sequences for a target domain, thereby enhancing the performance of a downstream HAR classifier. Let a skeletal motion sequence be defined as $\mathbf{x} = \{x(j) \in \mathbb{R}^d\}_{j=0}^{n-1}$, where $n$ is the number of frames and $d$ is the dimensionality of the pose representation. We consider two distinct domains:
\begin{enumerate}
    \item A prior domain $P$, characterized by a large-scale dataset $\mathcal{D}^P$, which is a collection of pairs $(\mathbf{x}^P, w)$. Here, $\mathbf{x}^P$ represents a motion sequence, and $w$ is its associated rich, free-form, descriptive text caption, from a set $W$.
    \item A target domain $T$, defined by a dataset $\mathcal{D}^T$, which is a collection of pairs $(\mathbf{x}^T, y)$. Here, $\mathbf{x}^T$ represents a motion sequence, and $y$ is its associated discrete action label from a set of action classes $Y$.
\end{enumerate}
Working in a few-shot setting implies that only a small subset of the target domain $\bar{T} \subset T$ is available at training time. Our goal is to use the limited set $\mathcal{D}^{\bar{T}}$ to adapt the generative model $G^P$, pre-trained on $\mathcal{D}^P$, yielding a new model $G^T$; capable of synthesizing novel, class-conditional motion samples from the action set $Y$. The quality of synthetic samples is measured by their ability to improve classification accuracy on $\mathcal{D}^T$ test split, when used for data augmentation in a HAR classifier training. The core challenge lies in bridging the significant domain gap between $\mathcal{D}^P$ and $\mathcal{D}^T$, which manifests in two ways:
\begin{enumerate}
    \item Conditioning modalities differ fundamentally (semantic gap). The source domain uses high-variance, free-form text, while the target domain employs discrete action labels.
    \item There is a discrepancy in the motion distributions (kinematic gap). Target motions tend to be more atomic and short, contrasting with the generally longer motions in the source domain. Furthermore, differences in data acquisition methods may contribute to further disparity.
\end{enumerate}

\begin{figure}[!t]
    \centering
    \includegraphics[width=0.75\columnwidth]{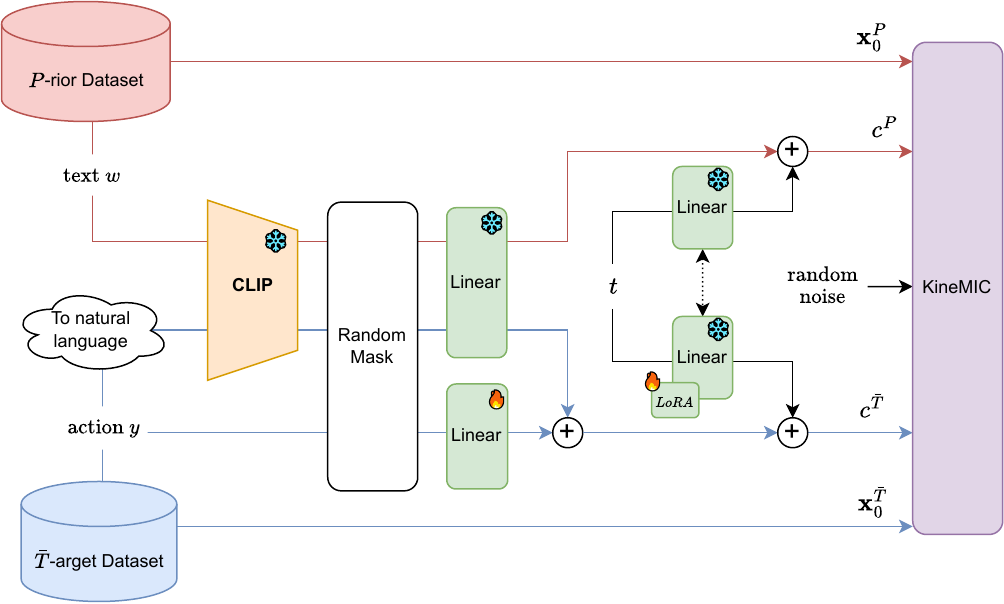}
    \caption{\textbf{Prior stream (red).} Conditioning signal $c^P$ uses CLIP text embedding and timestep. \textbf{Target stream (blue).} Conditioning signal $c^T$ uses action-label text embedding, learnable action embeddings and timestep. Dotted lines denote weight sharing.}
    \label{fig:KineMIC-inputs}
\end{figure}

\section{Methodology}
\label{sec:methodology}

\begin{figure}[!t]
    \centering
    \resizebox{1.0\textwidth}{!}{\includegraphics{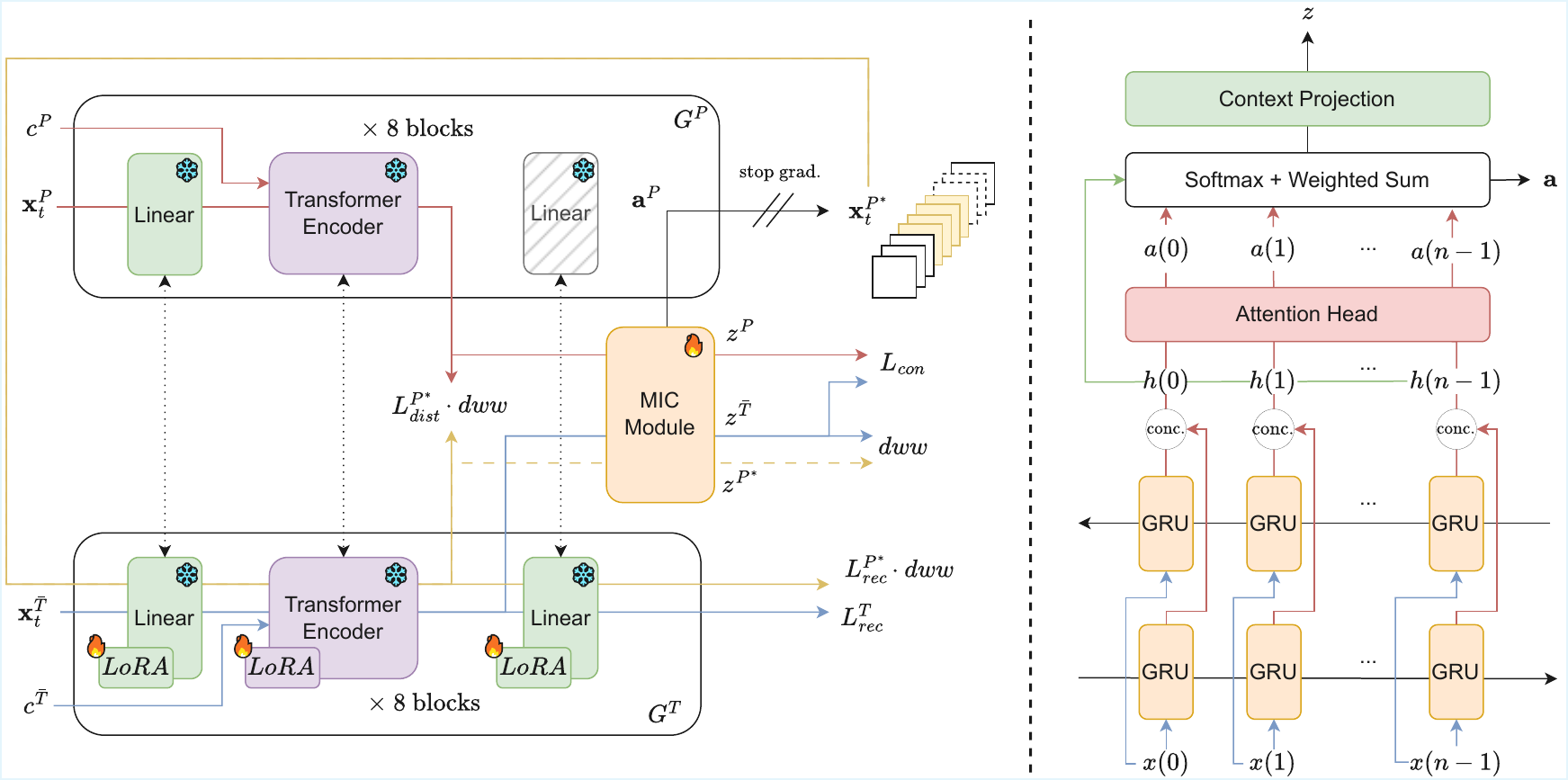}}
    \caption{\textbf{(Left) Core pipeline.} Dashed lines show detached gradients; dotted lines show weight sharing. MIC receives frame-wise tokens from teacher ($G^P$) and student ($G^T$) streams, producing latents $z^P$ and $z^{\bar{T}}$ for contrastive alignment and window mining. \textbf{(Right) MIC module.} attention-enhanced biGRU \cite{bahdanau2016neuralmachinetranslationjointly,gru} encoder aggregating frame-wise motion tokens into context-aware latents.}
    \label{fig:KineMIC}
\end{figure}

We introduce KineMIC (Figs.~\ref{fig:KineMIC-inputs} and~\ref{fig:KineMIC}), a unified framework integrating: semantic retrieval, kinematic alignment, kinematic mining, and student adaptation. These are detailed below, following the pipeline's information flow.

\subsection{Teacher-Student Architecture}

KineMIC builds on two identical MDM~\cite{tevet2022humanmotiondiffusionmodel} models: a frozen teacher $G^P$ and a trainable student $G^T$. Both models retain MDM's core structure and diffusion paradigm: predicting clean motion $\mathbf{x}_0$ (i.e. $\mathbf{x}$) from noised input $\mathbf{x}_t$ given $c$ conditioning signal. We refer to~\cite{tevet2022humanmotiondiffusionmodel} for full diffusion process details. To preserve prior knowledge, $G^T$ is fine-tuned with Low-Rank Adaptation (LoRA) \cite{hu2021loralowrankadaptationlarge}, which updates only a small fraction of weights through low-rank matrices while keeping most of parameters frozen. Furthermore, while the student inherits the frozen text embedding layer used in T2M, we introduce a learnable action embedding (Fig.~\ref{fig:KineMIC-inputs}) that enables $G^T$ specialization to discrete HAR labels.

\subsection{Semantic Alignment via Soft Positive Search}
Target labels in $\mathcal{D}^T$ lack the descriptive richness of T2M captions in $\mathcal{D}^P$, creating a one-to-many mapping (single label $\rightarrow$ multiple text prompts). We bridge this via CLIP-based \cite{radford2021learningtransferablevisualmodels} semantic retrieval. For each $y$ action label:
\begin{enumerate}
    \item Turn $y$ into natural-language (e.g. \textit{'side kick'} $\rightarrow$ \textit{'a person does a side kick'}).
    \item Encode both the target prompt and all captions in $\mathcal{D}^P$.
    \item Compute pairwise cosine similarities and retain top-$k$ closest samples.
\end{enumerate}
The resulting set of retrieved captions defines the \emph{soft positive matches} for each target label. This exploits CLIP's semantic space to correlate concepts even without explicit keyword overlap, modeling the one-to-many label-to-prompt mapping by identifying multiple semantically compatible descriptions that correspond to distinct yet potentially related motion patterns.

\subsection{Kinematic Alignment via MIC and Contrastive Learning}
Semantic similarity does not guarantee kinematic relevance. Moreover, $\mathcal{D}^P$ motions, designed for motion understanding through natural language, span longer time periods than atomic $\mathcal{D}^T$ actions designed for activity recognition, creating a structural disparity (\textit{see supplementary material for sequence-length analysis}). We address both via the Mining In Context (MIC) module, which learns frame-level alignments for targeted window extraction.

\subsubsection{MIC processing.} For each soft pair of two motion  sequences  $(\mathbf{x}^P, \mathbf{x}^{\bar{T}})$:
\begin{enumerate}
    \item A random timestep $t$ is sampled, and the forward diffusion process produces noised sequences $\mathbf{x}^P_t$ and $\mathbf{x}^{\bar T}_t$, starting from the same random noise.
    \item These are processed by $G^P$ and $G^T$ to obtain frame-wise feature tokens.
    \item Tokens from both streams are fed into the MIC module, a bidirectional GRU with attention \cite{bahdanau2016neuralmachinetranslationjointly,gru}, producing context-aware latent vectors $z^P$ and $z^{\bar T}$.
\end{enumerate}
These latent representations summarize the motion dynamics of each sample in a manner sensitive to both temporal structure and global context.

\subsubsection{Contrastive Alignment.}
MIC is trained to align $z^{\bar{T}_i}$ with $z^{P_j}$ from soft-positive pairs sharing the same target class $y$. Given a training batch $B = \{1,\ldots,b\}$ of size $b$, we use the Soft Nearest Neighbors loss \cite{frosst2019analyzingimprovingrepresentationssoft,pmlr-v2-salakhutdinov07a}, which naturally accommodates the one-to-many soft positives:
\begin{equation}
L_{con} = - \frac{1}{b}\sum_{i \in B}
\log \frac{
\sum_{j \in B^{+}(i)}
\exp(\text{sim}(z^{\bar{T}_i}, z^{P_j}) / \tau)
}{
\sum_{k \in B\setminus\{i\}}
\exp(\text{sim}(z^{\bar{T}_i}, z^{P_k}) / \tau)
}
\label{eq:contrastive}
\end{equation}
where $B^{+}(i)$ contains indices $j \in B \!\setminus\! \{i\}$ whose $z^{P_j}$ was retrieved as a soft positive for the same class $y$ as $z^{\bar{T}_i}$. This enforces kinematic alignment via a contrastive objective where $\text{sim}(\cdot, \cdot)$ denotes cosine similarity and $\tau$ temperature. This process allows MIC's attention to enable subsequent window extraction.

\subsection{Kinematic Mining}
With MIC trained via contrastive loss, its attention mechanism learns to focus on kinematically relevant frames in $\mathbf{x}^P$ to minimize the enforced alignment loss with $\mathbf{x}^{\bar{T}}$. Let $\mathbf{a}^P = \{a^P(k)\}_{k=0}^{n-1}$ denote attention weights over $\mathbf{x}^P$ of length $n$, and let $m$ be the length of $\mathbf{x}^{\bar{T}}$. We extract the most relevant contiguous subsequence (the "prior window") as the $m$-frame segment maximizing cumulative attention:
\begin{equation}
\mathbf{x}^{P^*}_t = \underset{ \{x^P_t(i),\ldots, x^P_t(i+m-1)\} \subseteq \mathbf{x}^P_t }{\arg\max} \sum_{k=i}^{i+m-1} a^P(k)
\end{equation}
The identified prior window $\mathbf{x}^{P^*}_t$ serves as a pseudo-labeled training sample for $G^T$ under the same target conditioning $c^T$ (Fig.~\ref{fig:KineMIC}).

\subsection{Optimization of Multi-Objective Loss Function}

\subsubsection{Reconstruction Loss.}
The student $G^T$ is trained to reconstruct both target motion and the extracted prior window:
\begin{equation}
L_{rec}^{T}
= \| \mathbf{x}^{\bar T}_0 - G^T(\mathbf{x}^{\bar T}_t, c^{\bar T}) \|_2^2
\label{eq:target_rec}
\end{equation}
\begin{equation}
L_{rec}^{P^*} = \| \mathbf{x}^{P^*}_0 - G^T(\mathbf{x}^{P^*}_t, c^{\bar T}) \|_2^2
\label{eq:window_rec}
\end{equation}
These losses ensure that $G^T$ learns both target-specific dynamics and source-domain motion structures consistent with the target action. Notice that both samples are processed enforcing the same $c^{\bar{T}}$ conditioning signal.

\subsubsection{Window Distillation.}
We distill the teacher’s internal representations relative to the mined window into the student. Let $\mathbf{u}^P$ and $\mathbf{u}^{P^*}$ denote pre-projection feature sequences of $\mathbf{x}^P_t$ and $\mathbf{x}^{P^*}_t$ processed by $G^P$ and $G^T$, respectively. For a prior window of length $m$ starting at frame index $i$, the distillation loss is:
\begin{equation}
L_{dist}^{P^*} = \frac{1}{m} \sum_{j=0}^{m-1} \| u^{P^*}(j) - u^P(i+j) \|_2^2
\label{eq:window_distill}
\end{equation}

\subsubsection{Dynamic Window Weighting.}
Mining is not perfect: a caption containing \textit{'kick'} word may retrieve \textit{'front-kick'} samples for a \textit{'side-kick'} target, yielding suboptimal kinematic windows despite semantic correlation. To modulate their influence, we compute a window quality score:
\begin{equation}
dww = (1 + \text{sim}(z^{\bar T}, z^{P^*}))/2
\label{eq:dww}
\end{equation}
where $\text{sim}(\cdot, \cdot)$ denotes cosine similarity and $z^{P^*}$ is computed with gradients detached, ensuring MIC remains trained exclusively by Eq.~\ref{eq:contrastive}. The score $dww$ scales both reconstruction (Eq.~\ref{eq:window_rec}) and distillation (Eq.~\ref{eq:window_distill}) losses, giving higher weight to more reliable prior windows.

\subsubsection{Complete Multi-Objective Loss Function.}
Final optimization follows a batch-averaged weighted objective:
\begin{equation}
L=\lambda_{rec}^T L_{rec}^T + \lambda_{con} L_{con} + dww \cdot (\lambda_{rec}^{P^*} L_{rec}^{P^*}+\lambda_{dist} L_{dist}^{P^*} )
\end{equation}

\section{Experiments}

\begin{figure}[t]
    \centering
    \includegraphics[width=1.0\linewidth]{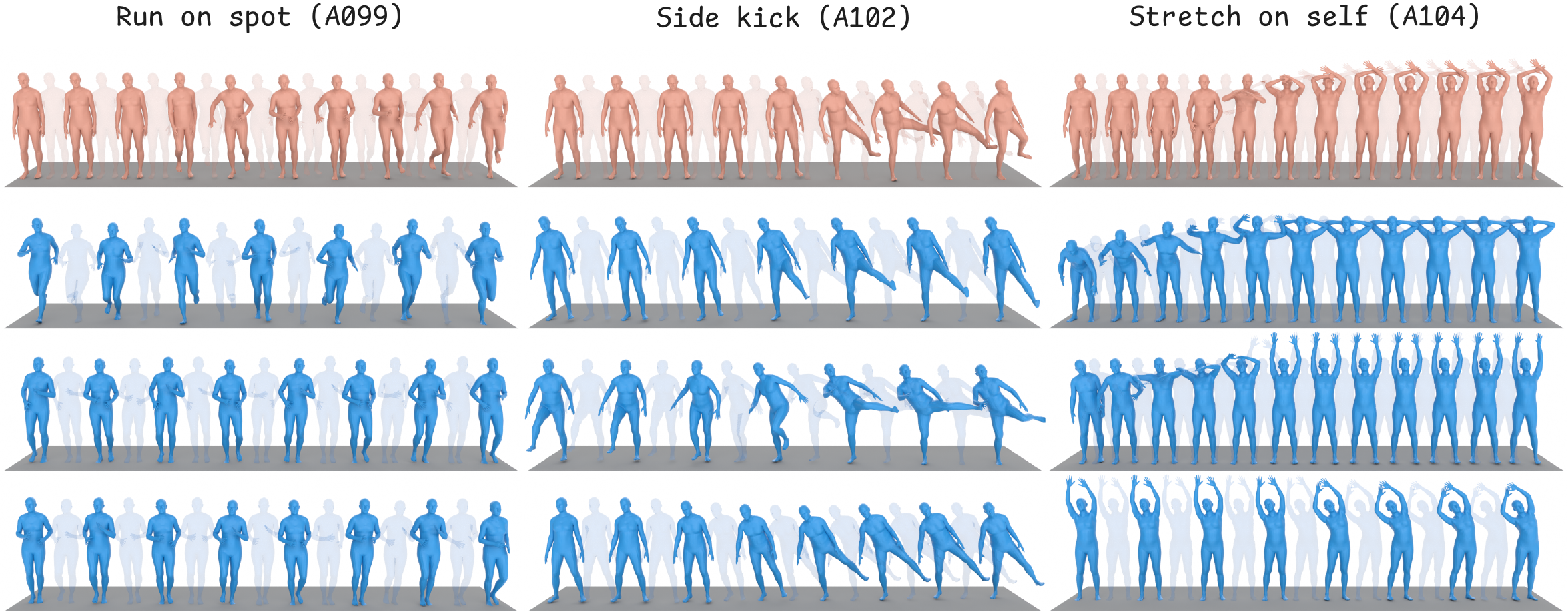}
    \caption{
        \textbf{Qualitative Examples.} Columns depict three target action classes. The first row shows corresponding real ($\mathcal{D}^{\bar{T}}$) ground truth samples from NTU RGB+D 120 extracted through VIBE~\cite{vibe} (in orange), while following rows show diverse samples generated by our KineMIC model (in blue). 
    }
    \label{fig:qualitative_results}
\end{figure}

\subsection{Experimental Setup}
Given the narrow scope of few-shot skeleton-based action synthesis for HAR, we position ourselves in the same experimental setup proposed by~\cite{fukushi2024fewshot} to contextualize our results. We define our source pre-train dataset ($\mathcal{D}^P$) as HumanML3D~\cite{t2m_humanml} and the target ($\mathcal{D}^T$) as a subset of NTU RGB+D 120~\cite{ntu120}, analyzing actions: \textit{'running on spot'} (A099), \textit{'side kick'} (A102), and \textit{'stretch on self'} (A104) from cross subject benchmark. Given the simplicity of these actions, generalist T2M models can reasonably generate them from broad text prompts. However, this setting precisely tests the challenge of capturing fine, class-discriminative kinematics critical for HAR. We choose these actions to evaluate KineMIC's ability to close this kinematic specificity gap. Following the few-shot protocol, we randomly select only 10 samples per action class (30 total) as our support set ($\mathcal{D}^{\bar{T}}$) for framework training. \textit{Further details in supplementary material.}

\subsubsection{Dataset and preprocessing.}
Following~\cite{fukushi2024fewshot}, 3D skeletons are re-estimated with VIBE~\cite{vibe}. To align NTU data with the HumanML3D pretraining~\cite{t2m_humanml} we apply two preprocessing steps: (1) drop hand joints from SMPL skeletons to match joint topology; (2) downsample sequences from 30 to 20 fps. All skeletons follow the HumanML3D normalization pipeline: root centering, feet height set to zero, initial pose oriented to +Z, and joint-length normalization across frames. Motions are represented with the 263‑dim vector used in MDM~\cite{tevet2022humanmotiondiffusionmodel}, concatenating root-relative joint positions, 6D joint rotations, joint velocities, and binary foot‑contact flags.

\begin{figure}[t]
    \centering
    \includegraphics[width=1.0\linewidth]{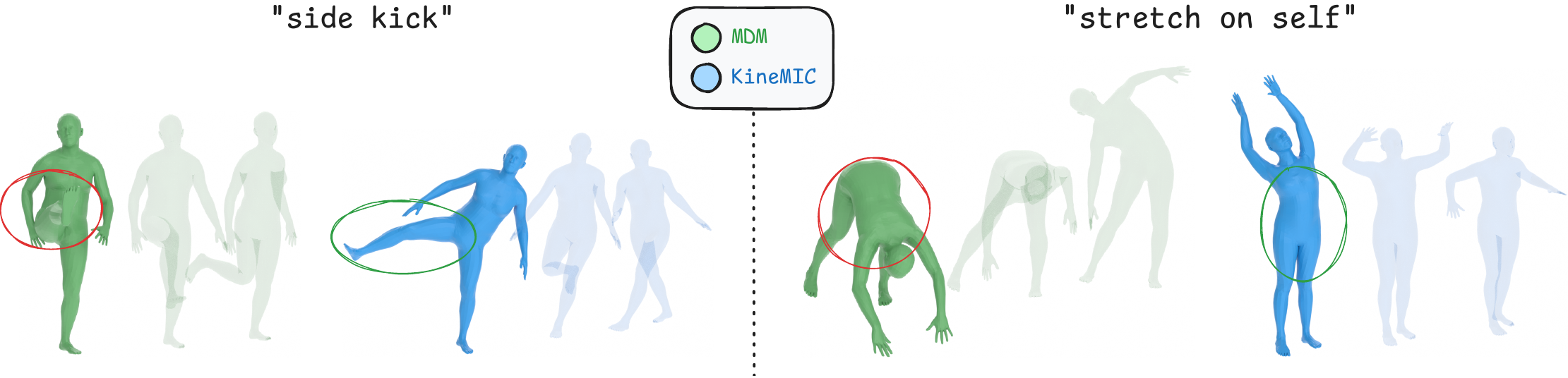}
    \caption{\textbf{Qualitative Comparison.} We compare pre-trained MDM (green) against KineMIC (blue). MDM incorrectly produces a \textit{'front kick'} for the \textit{'side kick'} prompt, whereas KineMIC captures the correct leg orientation. For \textit{'stretch on self'}, MDM generates a generic \textit{'pike'} pose; while semantically valid for \textit{'stretching'}, this deviates from the standing poses observed in $\mathcal{D}^{\bar{T}}$ (Fig.~\ref{fig:qualitative_results}) which KineMIC correctly recovers.
    }
    \label{fig:comparison}
\end{figure}

\subsubsection{Implementation Details.}
The pre-trained $G^P$ is an 8-blocks MDM  transformer encoder~\cite{tevet2022humanmotiondiffusionmodel} trained on HumanML3D~\cite{t2m_humanml}. $G^T$ inherits the architecture and weights of $G^P$, adding only an action embedding layer (Fig.~\ref{fig:KineMIC-inputs}) and the MIC module. We fine-tune $G^T$ using LoRA~\cite{hu2021loralowrankadaptationlarge} (rank=16, $\alpha=32$, dropout=0.1) on all transformer layers, optimizing: the low-rank matrices, the MIC module and the action embedding. All $\lambda$ parameters are set to $1.0$. For soft positive search, we set $k=250$ to balance diversity and semantic relevance, with temperature $\tau=0.07$. Soft positive matches can be computed once before training, yielding an effective training dataset size of $(\textit{Shots} + k) \times \textit{Classes}$. We employ classifier-free guidance (scale 2.5)~\cite{tevet2022humanmotiondiffusionmodel}, dropping action and text modalities each with 0.1 probability. Models are trained for 5000 steps (AdamW, LR $2 \cdot 10^{-5}$) using 100 diffusion steps, cosine noise scheduling, and gradient clipping (norm 1.0). Each training step processes all $\mathcal{D}^{\bar{T}}$ samples (30 total), pairing each $(\mathbf{x}^{\bar{T}}, y)$ with a random soft-pair $(\mathbf{x}^P, w)$ from the top-$k$ matches in $\mathcal{D}^P$ for class $y$.

\begin{figure}[t]
    \centering
    \includegraphics[width=1.0\linewidth]{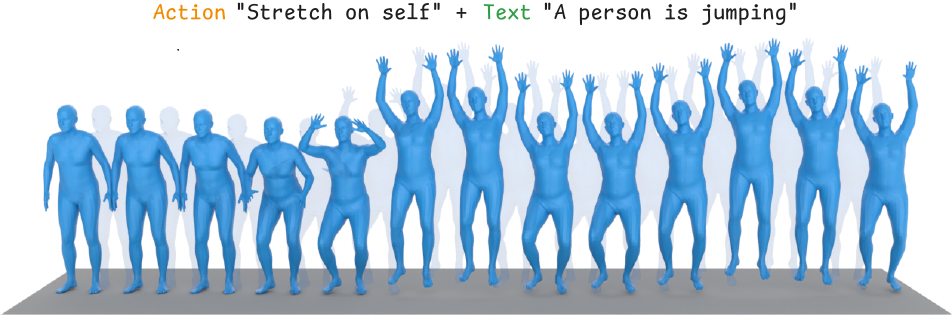}
    \caption{\textbf{Motion composition via dual conditioning.} Given two unrelated prompts (action and text), KineMIC generates meaningful, novel motion that combines both instructions without explicit training on this task.
    }
    \label{fig:composition}
\end{figure}

\subsubsection{Evaluation Protocol.}
Our evaluation focuses on maximizing synthetic data utility for HAR, centered on recognition accuracy, Diversity (motion distribution variability) and MultiModality (per-prompt variance). We employ a ST-GCN classifier~\cite{yan2018spatialtemporalgraphconvolutional} for all downstream HAR and generative evaluations, following PYSKL~\cite{duan2022pysklgoodpracticesskeleton} practices and implementation. Each training set is composed of 30 real samples augmented with 1152 synthetically generated motions, both uniformly distributed per class. Tests denoting "real-data-only" refer to experiments using only real samples ($\mathcal{D}^{\bar{T}}$) without synthetic augmentation. All experiments are repeated five times with different seeds. For comparison with~\cite{fukushi2024fewshot}, we report the median top-1 accuracy; internal analyses use mean and standard deviation.

\subsection{Baselines and Prior Work Comparison}
\begin{figure}
\centering
\includegraphics[width=0.85\textwidth]{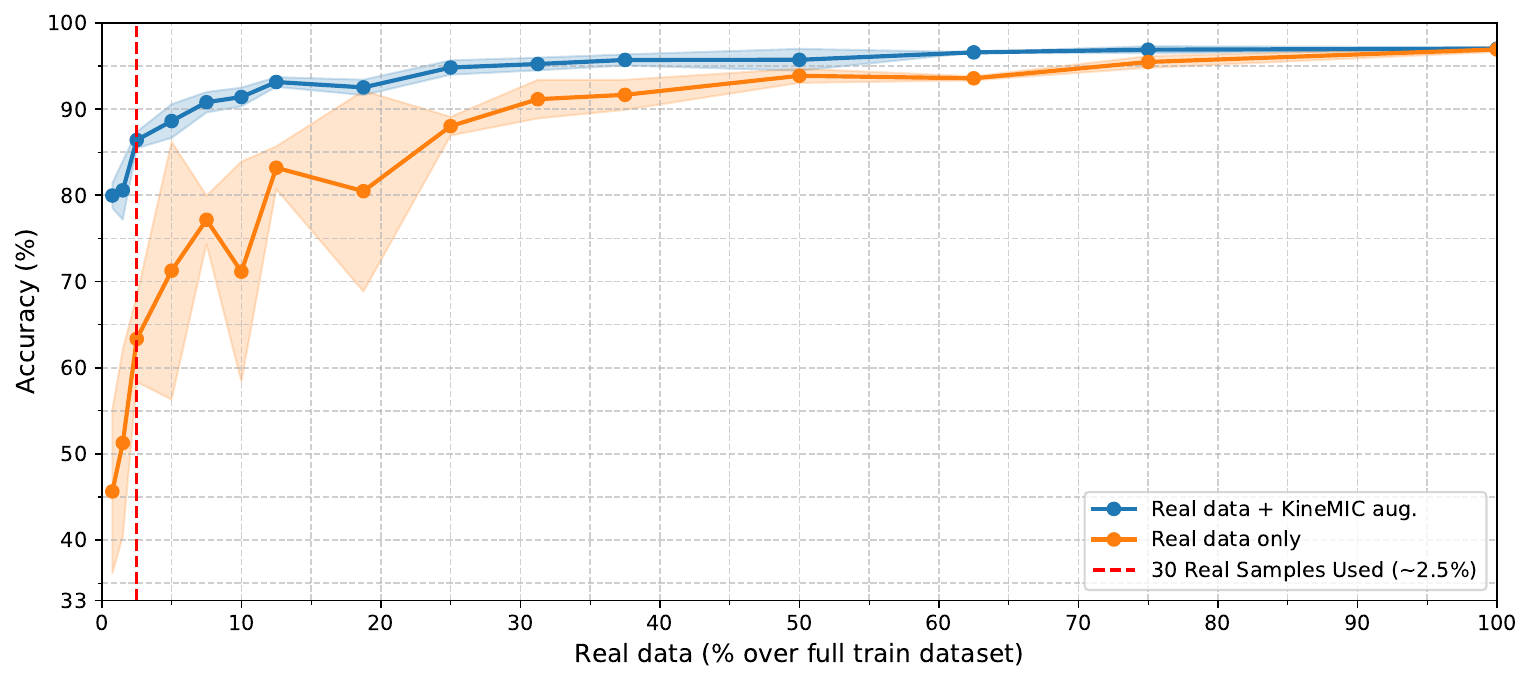}
\caption{\textbf{Synthetic data scalability.} When the ST-GCN classifier is trained using only real $\mathcal{D}^T$ data, training exhibits high variance and poor performances. When synthetic data from KineMIC is concatenated, accuracy is consistently much higher.} \label{fig:incremental}
\end{figure}

In Table~\ref{tab:prior_work_comparison} we compare the KineMIC framework against key prior work and critical baselines. First, we establish a foundational baseline using the pre-trained MDM model by generating motions from general text prompts describing our target classes (e.g., converting class \textit{'side kick'} to \textit{'a person does a side kick'}), without any fine-tuning. This achieves 83.9\%, confirming that the generalist MDM already possesses substantial motion knowledge relevant to our chosen actions and providing a strong starting point. Training MDM from random initialization (73.9\%) or fine-tuning from a $\mathcal{D}^P$ checkpoint with LoRA (75.5\%) both result in significantly lower performance. We attribute this to severe overfitting on the extremely limited few-shot data, causing the diffusion model to collapse. In stark contrast, KineMIC better leverages pre-training knowledge while preventing overfitting, achieving 86.2\%. Our best results match \cite{fukushi2024fewshot}, but through a substantially different approach that leverages T2M diffusion priors rather than GAN-based regularization. Relative to real-data-only training on $\mathcal{D}^{\bar{T}}$ samples (63.1\%), KineMIC delivers a substantial +23.1\% point improvement. Figure~\ref{fig:incremental} demonstrates scalability: real-data-only training shows poor performance and high variance in low-data regimes, while KineMIC augmentation consistently improves accuracy and asymptotically approaches full $\mathcal{D}^T$ training performance.

\begin{table}[t]
\centering
\caption{
    \textbf{Comparative Performance on Few-Shot HAR.} Median top-1 accuracy on the NTU RGB+D 120 dataset, following the evaluation protocol defined by prior work of Fukushi et al.~\cite{fukushi2024fewshot}. Results marked with $\dag$ are reported directly from~\cite{fukushi2024fewshot}. The up-arrow ($\uparrow$) indicates that higher is better.}
\label{tab:prior_work_comparison}
\begin{tabular}{|l|l|c|} 
\hline
\textbf{Source} & \textbf{Method} & {\textbf{Top-1 Acc (\%)} $\uparrow$} \\
\hline
\multirow{4}{*}{Prior Work$^\dag$} & Real-data-only (30 samples)$^\dag$ & 58.4 \\
& ACTOR$^\dag$~\cite{petrovich2021actionconditioned3dhumanmotion} & 73.6 \\
& Kinetic-GAN$^\dag$~\cite{degardin2021generativeadversarialgraphconvolutional} & 81.7 \\
& Fukushi et al.$^\dag$~\cite{fukushi2024fewshot} & \underline{86.4} \\
\hline 
\multirow{5}{*}{Our Analysis} & Real-data-only (30 samples) & 63.1 \\
& MDM (pre-trained) & 83.9 \\
& MDM (from scratch) & 73.9 \\
& MDM (LoRA fine-tune) & 75.5 \\
& KineMIC & 86.2 \\
\hline 
\end{tabular}
\end{table}

\subsection{Ablation Study}
In Table~\ref{tab:ablations} we conduct a systematic ablation study to evaluate the contribution of each component within our framework. Starting from baselines, the pre-trained MDM shows highest diversity (26.82) but largest error bounds (85.23$\pm$3.32\%), as its generic samples deviate from desirable $\mathcal{D}^T$ kinematics (Fig.~\ref{fig:comparison}) despite visual variety, producing noisy training signals for the downstream classifier. In contrast, our LoRA fine-tuning baseline overfits on the extremely limited few-shot data, collapsing diversity to 9.45. Introducing KineMIC components, our base model (contrastive alignment + reconstruction) matches pre-trained accuracy while reducing variance (85.21$\pm$2.24\%). Interestingly, distilling teacher features ($L_{dist}$) actually reduces accuracy (83.78\%) despite achieving peak multimodality (17.12). This suggests $L_{dist}$ enforces the teacher's generalist T2M behavior rather than promoting the HAR-specific kinematic consistency we seek. Ultimately, enabling dynamic window weighting ($dww$) alone produces the best, most consistent results (86.41$\pm$0.95\%). This strongly supports our core hypothesis: while semantic similarity provides valuable guide toward kinematic alignment, it is not always reliable and requires intelligent quality filtering to deliver stable HAR performance.
\begin{table}[t]
\centering
\caption{
    \textbf{Ablation Study.} We evaluate the contribution of each component to our framework. Baselines are pre-trained MDM and LoRA fine-tuned MDM. Our core KineMIC (Base) model builds on this by adding Eq. \ref{eq:contrastive} and \ref{eq:target_rec} respectively. We then incrementally add $L_{\text{dist}}$ and $dww$. ($\uparrow$) denotes that higher values are better.
}
\label{tab:ablations}
\small 
\renewcommand{\arraystretch}{1.1} 
\begin{tabular}{|l|c|c|c|} 
\hline 
\textbf{Method} & \textbf{Accuracy (\%) $\uparrow$} & \textbf{Diversity $\uparrow$} & \textbf{MultiModality $\uparrow$}\\
\hline
MDM (pre-trained) & $85.23^{\pm 3.32}$ & $\underline{26.82^{\pm 1.20}}$ & $6.10^{\pm 2.42}$ \\
MDM (LoRA fine-tune) & $76.27^{\pm 1.51}$ & $9.45^{\pm 0.39}$ & $4.98^{\pm 0.36}$ \\
\hline
KineMIC (Base) & $85.21^{\pm 2.24}$ & $13.69^{\pm 1.10}$ & $10.21^{\pm 0.60}$ \\
+ $L_{\text{dist}}$ only & $83.78^{\pm 2.38}$ & $22.45^{\pm 1.56}$ & \underline{$17.12^{\pm 2.25}$} \\
+ $dww$ only & \underline{$86.41^{\pm 0.95}$} & $14.24^{\pm 1.09}$ & $10.58^{\pm 0.77}$ \\
+ $L_{\text{dist}}$ + $dww$ & $84.94^{\pm 2.37}$ & $19.49^{\pm 1.30}$ & $13.64^{\pm 0.91}$ \\
\hline
\end{tabular}
\end{table}

\subsubsection{Training and Inference Cost.}
Training is efficient due to the few-shot design, requiring only 10 samples per class. While our best model consists of $\sim$35M parameters, the application of LoRA \cite{hu2021loralowrankadaptationlarge} reduces the trainable parameters to $\sim$0.9M (3\% of the total). A full training session requires 2–3 hours on a commercial NVIDIA T4 GPU, and inference for a 50-frame motion takes $\sim$2 seconds. Furthermore, unlike \cite{fukushi2024fewshot}, our approach trains a single unified model for all classes.

\subsection{Qualitative Evaluation}
We conduct a comprehensive visual assessment of 90 generated samples from our best KineMIC model, alongside 90 outputs from the pre-trained MDM baseline and all 30 real $\mathcal{D}^{\bar{T}}$ samples (\textit{see supplementary material}). Figure~\ref{fig:qualitative_results} presents few examples, demonstrating KineMIC's ability to produce motions coherent with target kinematics while exhibiting rich, semantically meaningful variations, as supported by ablation gains in accuracy and diversity (Tab.~\ref{tab:ablations}). For the \textit{'side kick'} class, KineMIC generates motions ranging from $\mathcal{D}^{\bar{T}}$ replicas to dynamic combat-style kicks, demonstrating nuanced understanding beyond the sparse $\mathcal{D}^{\bar{T}}$ data. The \textit{'stretch on self'} action produces enriched variations including stretches with hands behind the head or lateral torso tilts, while \textit{'run on spot'} yields natural pacing and arm swing differences. As shown in Figure~\ref{fig:comparison}, while pre-trained MDM generates motions that are semantically relevant to the input text, it does not always capture the precise kinematic fidelity required for HAR. KineMIC improves upon this by aligning generation with $\mathcal{D}^{\bar{T}}$ requirements, ensuring the accurate synthesis of movements such as directional kicks or specific stretching postures that may not be recovered by the baseline.

\subsubsection{Motion Composition.}
KineMIC exhibits emergent compositional behavior when guided by text prompts differing from the action-conditioning class. For instance, conditioning on the \textit{'stretch on self'} action ID while prompting with \textit{'a person is jumping'} generates a figure \textit{'jumping while stretching arms overhead'} (Fig.~\ref{fig:composition}). This property is most effective when controls are kinematically non-conflicting (e.g., combining upper-body dominant actions with lower-body ones), whereas conflicting lower-body prompts can lead to instability. This ability emerges from a synergy between our dual-conditioning scheme (Fig.~\ref{fig:KineMIC-inputs}) and LoRA, which preserves pre-training capabilities during fine-tuning~\cite{sawdayee2025dancelikechickenlowrank}.
\section{Conclusions}

This work introduces and addresses the challenge of few-shot action synthesis for HAR using T2M priors. We show that while T2M models provide a strong baseline, they lack the kinematic specificity for specialized action classes. To resolve this, KineMIC adapts a pre-trained diffusion model using only 10 samples per class via a soft positive search in a shared semantic space and a kinematic mining strategy. Empirically, KineMIC improves downstream classifier performance on the NTU-120 benchmark \cite{fukushi2024fewshot}. While naive fine-tuning leads to collapse (73.9–75.5\%), KineMIC's context-aware mining delivers stable +23.1\% point gains over real-data-only baselines (63.1\% $\rightarrow$ 86.2\%), showing that T2M priors are viable for few-shot HAR.

\subsubsection{Limitations and future directions.} The primary constraint lies in the core assumption that semantic correspondence is a reliable proxy for kinematic relevance during the mining process. This assumption does not always hold; for instance, text encoders may assign high similarity to \textit{`punch'} and \textit{`kick'} due to shared concepts such as \textit{`fighting'}, despite their substantial kinematic disparity, or struggle with generic concepts such as \textit{'stretching'}. This increases the importance of effective filtering through mechanisms like the \textit{dww} we proposed. Additionally, robustness is intrinsically linked to (a) the scale and diversity of the T2M source data and (b) the specificity of the target actions, implying a practical limit when mining highly complex and specific movements. Consequently, expanding the framework to encompass larger and more complex action sets remains a key direction for future work to further enhance its generalization. Furthermore, prompt augmentation strategies were not explored. Integrating such techniques is expected to enhance the semantic richness of the target conditioning, a critical factor for fully exploiting the potential of T2M priors.

\subsubsection{Declarations.}
\textit{Use of LLMs:} AI tools were used solely for language refinement and technical editing. The core scientific contributions are the work of the authors, who hold full responsibility for the paper's content.
\textit{Acknowledgments:} This work was conducted by the first author during an internship period at CESI LINEACT under the second author's supervision.

%
%

\bibliographystyle{splncs04}
\bibliography{biblio}

@inproceedings{bahdanau2016neuralmachinetranslationjointly,
    author = {Bahdanau, D. and Cho, K. and Bengio, Y.},
    title = {Neural Machine Translation by Jointly Learning to Align and Translate},
    booktitle = {ICLR},
    year = {2015},
    doi = {10.48550/arXiv.1409.0473},
}

@INPROCEEDINGS{chi2022infogcn,
    author = {Chi, Hyung-Gun and Ha, Myoung Hoon and Chi, Seunggeun and Lee, Sang Wan and Huang, Qixing and Ramani, Karthik},
    title = {Info{GCN}: Representation Learning for Human Skeleton-based Action Recognition}, 
    booktitle = {CVPR},
    year = {2022},
    doi = {10.1109/CVPR52688.2022.01955},
}

@inproceedings{gru,
    author = {Junyoung Chung and Caglar Gulcehre and KyungHyun Cho and Yoshua Bengio},
    title = {Empirical Evaluation of Gated Recurrent Neural Networks on Sequence Modeling},
    year = {2014},
    booktitle = {NIPS Workshop},
    doi = {10.48550/arXiv.1412.3555},
}

@inproceedings{degardin2021generativeadversarialgraphconvolutional,
    author = {Bruno Degardin and Joao Neves and Vasco Lopes and Joao Brito and Ehsan Yaghoubi and Hugo Proenca},
    title = {Generative Adversarial Graph Convolutional Networks for Human Action Synthesis}, 
    booktitle = {WACV},
    year = {2022},
    doi = {10.1109/WACV51458.2022.00281},
}

@article{Dentamaro2024HumanAR,
    author = {Vincenzo Dentamaro and Vincenzo Gattulli and Donato Impedovo and Fabio Manca},
    title = {Human activity recognition with smartphone-integrated sensors: A survey},
    journal = {Expert Syst. Appl.},
    year = {2024},
    doi = {10.1016/J.ESWA.2024.123143},
}

@inproceedings{do2024skateformerskeletaltemporaltransformerhuman,
    author={Jeonghyeok Do and Munchurl Kim},
    title={SkateFormer: Skeletal-Temporal Transformer for Human Action Recognition}, 
    booktitle = {ECCV},
    year = {2024},
    doi = {10.1007/978-3-031-72940-9\_23},
}

@INPROCEEDINGS{du2015hierarchical,
    author = {Yong Du and Wang, Wei and Wang, Liang},
    title = {Hierarchical recurrent neural network for skeleton based action recognition},
    booktitle = {CVPR}, 
    year = {2015},
    doi = {10.1109/CVPR.2015.7298714},
}

@inproceedings{duan2022pysklgoodpracticesskeleton,
    author = {Haodong Duan and Jiaqi Wang and Kai Chen and Dahua Lin},
    title = {{PYSKL}: Towards Good Practices for Skeleton Action Recognition}, 
    booktitle = {ACM MM},
    year = {2022},
    doi = {10.1145/3503161.3548546},
}

@inproceedings{duan2022revisitingskeletonbasedactionrecognition,
    author={Haodong Duan and Yue Zhao and Kai Chen and Dahua Lin and Bo Dai},
    title={Revisiting Skeleton-based Action Recognition}, 
    booktitle = {CVPR},
    year = {2022},
    doi = {10.1109/CVPR52688.2022.00298},
}

@inproceedings{frosst2019analyzingimprovingrepresentationssoft,
    author={Nicholas Frosst and Nicolas Papernot and Geoffrey Hinton},
    title={Analyzing and Improving Representations with the Soft Nearest Neighbor Loss}, 
    booktitle = {ICML},
    year = {2019},
    url = {http://proceedings.mlr.press/v97/frosst19a.html},
}

@INPROCEEDINGS{fukushi2024fewshot,
    author = {Fukushi, Kenichiro and Nozaki, Yoshitaka and Nishihara, Kosuke and Nakahara, Kentaro},
    title = {Few-shot generative model for skeleton-based human action synthesis using cross-domain adversarial learning}, 
    booktitle = {WACV},
    year = {2024},
    doi = {10.1109/WACV57701.2024.00390},
}

@inproceedings{guo2023momaskgenerativemaskedmodeling,
    author = {Chuan Guo and Yuxuan Mu and Muhammad Gohar Javed and Sen Wang and Li Cheng},
    title = {Mo{M}ask: Generative Masked Modeling of 3D Human Motions}, 
    booktitle = {CVPR},
    year = {2024},
    doi = {10.1109/CVPR52733.2024.00186},
}

@INPROCEEDINGS{t2m_humanml,
    author = {Guo, Chuan and Zou, Shihao and Zuo, Xinxin and Wang, Sen and Ji, Wei and Li, Xingyu and Cheng, Li},
    title = {Generating Diverse and Natural 3D Human Motions from Text}, 
    booktitle = {CVPR}, 
    year = {2022},
    doi = {10.1109/CVPR52688.2022.00509},
}

@inproceedings{Guo_2020,
    author = {Guo, Chuan and Zuo, Xinxin and Wang, Sen and Zou, Shihao and Sun, Qingyao and Deng, Annan and Gong, Minglun and Cheng, Li},
    title = {Action2{M}otion: Conditioned Generation of 3D Human Motions},
    booktitle = {ACM MM},
    year = {2020},
    doi = {10.1145/3394171.3413635},
}

@inproceedings{hu2021loralowrankadaptationlarge,
    author = {Edward J. Hu and Yelong Shen and Phillip Wallis and Zeyuan Allen-Zhu and Yuanzhi Li and Shean Wang and Lu Wang and Weizhu Chen},
    title = {Lo{RA}: Low-Rank Adaptation of Large Language Models}, 
    booktitle = {ICLR},
    year = {2022},
    url = {https://openreview.net/forum?id=nZeVKeeFYf9},
}

@article{HungCuong2023DeepLF,
    author = {Nguyen Hung-Cuong and Thi-Hao Nguyen and Rafal Scherer and Van-Hung Le},
    title = {Deep Learning for Human Activity Recognition on 3D Human Skeleton: Survey and Comparative Study},
    journal = {Sensors (Basel, Switzerland)},
    year = {2023},
    doi = {10.3390/S23115121},
}

@INPROCEEDINGS{vibe,
    author = {Kocabas, Muhammed and Athanasiou, Nikos and Black, Michael J.},
    title = {{VIBE}: Video Inference for Human Body Pose and Shape Estimation}, 
    booktitle = {CVPR},
    year = {2020},
    doi = {10.1109/CVPR42600.2020.00530},
}

@article{leng2025scalinghumanactivityrecognition,
    author={Zikang Leng and Archith Iyer and Thomas Plotz},
    title={Scaling Human Activity Recognition: A Comparative Evaluation of Synthetic Data Generation and Augmentation Techniques}, 
    journal = {arXiv},
    year = {2025},
    doi = {10.48550/ARXIV.2506.07612},
}

@article{ntu120,
    author = {Jun Liu and Amir Shahroudy and Mauricio Perez and Gang Wang and Ling{-}Yu Duan and Alex C. Kot},
    title = {{NTU} {RGB+D} 120: {A} Large-Scale Benchmark for 3D Human Activity Understanding},
    journal = {TPAMI},
    year = {2020},
    doi = {10.1109/TPAMI.2019.2916873},
}

@article{lupion2024dataaugmentation,
    author = "Marcos Lupion and Federico Cruciani and I Cleland and CD Nugent and Pilar Ortigosa",
    title = "Data augmentation for Human Activity Recognition with Generative Adversarial Networks",
    journal = {IEEE J. Biomed. Health Informatics},
    year = {2024},
    doi = {10.1109/JBHI.2024.3364910},
}

@inproceedings{mahmood2019amassarchivemotioncapture,
    author = {Naureen Mahmood and Nima Ghorbani and Nikolaus F. Troje and Gerard Pons-Moll and Michael J. Black},
    title = {{AMASS}: Archive of Motion Capture as Surface Shapes}, 
    booktitle = {ICCV},
    year = {2019},
    doi = {10.1109/ICCV.2019.00554},
}

@inproceedings{petrovich2021actionconditioned3dhumanmotion,
    author = {Mathis Petrovich and Michael J. Black and Gul Varol},
    title = {Action-Conditioned 3D Human Motion Synthesis with Transformer {VAE}}, 
    booktitle = {ICCV},
    year = {2021},
    doi = {10.1109/ICCV48922.2021.01080},
}

@inproceedings{petrovich2022temosgeneratingdiversehuman,
    author = {Mathis Petrovich and Michael J. Black and Gul Varol},
    title = {{TEMOS}: Generating diverse human motions from textual descriptions}, 
    booktitle = {ECCV},
    year = {2022},
    doi = {10.1007/978-3-031-20047-2\_28},
}

@inproceedings{plizzari2021spatio,
    author = {Plizzari, Chiara and Cannici, Marco and Matteucci, Matteo},
    title = {Spatial Temporal Transformer Network for Skeleton-Based Action Recognition},
    booktitle = {ICPR Workshop},
    year = {2021},
    doi = {10.1007/978-3-030-68796-0\_50},
}

@inproceedings{radford2021learningtransferablevisualmodels,
    author={Alec Radford and Jong Wook Kim and Chris Hallacy and Aditya Ramesh and Gabriel Goh and Sandhini Agarwal and Girish Sastry and Amanda Askell and Pamela Mishkin and Jack Clark and Gretchen Krueger and Ilya Sutskever},
    title={Learning Transferable Visual Models From Natural Language Supervision}, 
    booktitle = {ICML},
    year = {2021},
    url = {http://proceedings.mlr.press/v139/radford21a.html},
}

@inproceedings{pmlr-v2-salakhutdinov07a,
    author = {Salakhutdinov, Ruslan and Hinton, Geoff},
    title = {Learning a Nonlinear Embedding by Preserving Class Neighbourhood Structure},
    booktitle = {AISTATS},
    year = {2007},
    url = {http://proceedings.mlr.press/v2/salakhutdinov07a.html},
}

@article{sawdayee2025dancelikechickenlowrank,
    author = {Haim Sawdayee and Chuan Guo and Guy Tevet and Bing Zhou and Jian Wang and Amit H. Bermano},
    title = {Dance Like a Chicken: Low-Rank Stylization for Human Motion Diffusion}, 
    journal = {arXiv},
    year = {2025},
    doi = {10.48550/ARXIV.2503.19557},
}

@inproceedings{shafir2023humanmotiondiffusiongenerative,
    author={Yonatan Shafir and Guy Tevet and Roy Kapon and Amit H. Bermano},
    title={Human Motion Diffusion as a Generative Prior}, 
    booktitle = {ICLR},
    year = {2024},
    url = {https://openreview.net/forum?id=dTpbEdN9kr},
}

@inproceedings{ntu60,
    author = {Amir Shahroudy and Jun Liu and Tian-Tsong Ng and Gang Wang},
    title = {{NTU} {RGB+D:} A Large Scale Dataset for 3D Human Activity Analysis},
    booktitle = {CVPR},
    year = {2016},
    doi = {10.1109/CVPR.2016.115},
}

@inproceedings{shi2019twostreamadaptivegraphconvolutional,
    author = {Lei Shi and Yifan Zhang and Jian Cheng and Hanqing Lu},
    title = {Two-Stream Adaptive Graph Convolutional Networks for Skeleton-Based Action Recognition}, 
    booktitle = {CVPR},
    year = {2019},
    doi = {10.1109/CVPR.2019.01230},
}

@inproceedings{tevet2022humanmotiondiffusionmodel,
    author = {Guy Tevet and Sigal Raab and Brian Gordon and Yonatan Shafir and Daniel Cohen-Or and Amit H. Bermano},
    title = {Human Motion Diffusion Model}, 
    booktitle = {ICLR},
    year = {2023},
    url = {https://openreview.net/forum?id=SJ1kSyO2jwu}
}

@article{wanyan2025comprehensivereviewfewshotaction,
    author = {Yuyang Wanyan and Xiaoshan Yang and Weiming Dong and Changsheng Xu},
    title = {A Comprehensive Review of Few-shot Action Recognition}, 
    journal = {Int. J. Comput. Vis.},
    year = {2025},
    doi = {10.1007/S11263-025-02503-6},
}

@inproceedings{yan2018spatialtemporalgraphconvolutional,
    author = {Sijie Yan and Yuanjun Xiong and Dahua Lin},
    title = {Spatial Temporal Graph Convolutional Networks for Skeleton-Based Action Recognition}, 
    booktitle = {AAAI},
    year = {2018},
    doi = {10.1609/AAAI.V32I1.12328},
}

@article{zhang2025motionanythingmotiongeneration,
    author = {Zeyu Zhang and Yiran Wang and Wei Mao and Danning Li and Rui Zhao and Biao Wu and Zirui Song and Bohan Zhuang and Ian Reid and Richard Hartley},
    title = {Motion Anything: Any to Motion Generation}, 
    journal = {arXiv},
    year = {2025},
    doi = {10.48550/ARXIV.2503.06955},
}

\end{document}